\documentclass[lettersize,journal]{IEEEtran}
\usepackage{amsmath,amsfonts}
\usepackage{algorithmic}
\usepackage{algorithm}
\usepackage{array}
\usepackage{subfigure}
\usepackage{multirow}
\usepackage[caption=false,font=normalsize,labelfont=sf,textfont=sf]{subfig}
\usepackage{textcomp}
\usepackage{stfloats}
\usepackage{url}
\usepackage{verbatim}
\usepackage{graphicx}
\usepackage{cite}
\usepackage{pgfplots}
\usepackage{color,xcolor}
\usepackage{CJKutf8}
\usepackage{amsmath,amsfonts}
\usepackage{algorithmic}
\usepackage{array}
\usepackage[caption=false,font=normalsize,
   labelfont=sf,textfont=sf]{subfig}
\usepackage{textcomp}
\usepackage{stfloats}
\usepackage{url}
\usepackage{verbatim}
\usepackage{graphicx}
\usepackage{balance}
\usepackage{inconsolata}
\usepackage{booktabs}
\usepackage{paralist}
\usepackage{colortbl}
\usepackage{algorithm}
\usepackage{algorithmic}
\usepackage{makecell}
\usepackage{arydshln}
\usepackage{multirow}
\usepackage{color}
\usepackage{pgfplots}
\usepackage[normalem]{ulem}
\usepackage{graphicx}
\usepackage{amsfonts}
\usepackage{amsmath,amssymb,bm}
\usepackage{soul}

\begin{document}
\begin{CJK}{UTF8}{gbsn}

\title{ECQED: Emotion-Cause Quadruple\\ Extraction in Dialogs}

\author{Li Zheng, Donghong Ji, Fei Li, Hao Fei, Shengqiong Wu, Jingye Li, Bobo Li and Chong Teng
\thanks{Li Zheng, Donghong Ji, Fei Li, Jingye Li, Bobo Li and Chong Teng are with the Key Laboratory of Aerospace Information Security and Trusted Computing, Ministry of Education, School of Cyber Science and Engineering, Wuhan University, Wuhan 430072, China
(e-mail: zhengli@whu.edu.cn; dhji@whu.edu.cn; foxlf823@gmail.com; theodorelee@whu.edu.cn; 
boboli@whu.edu.cn;
tengchong@whu.edu.cn).
}
\thanks{Hao Fei and Shengqiong Wu are with the Sea-NExT Joint Lab, School of Computing, National University of Singapore, Brazil (e-mail: haofei37@nus.edu.sg; 
whuwsq@whu.edu.cn).}
}

\markboth{Journal of \LaTeX\ Class Files,~Vol.~14, No.~8, August~2021}%
{Shell \MakeLowercase{\textit{et al.}}: A Sample Article Using IEEEtran.cls for IEEE Journals}


\maketitle

\begin{abstract}
The existing emotion-cause pair extraction (ECPE) task, unfortunately, ignores extracting the \emph{emotion type} and \emph{cause type}, while these fine-grained meta-information can be practically useful in real-world applications, i.e., chat robots and empathic dialog generation.
Also the current ECPE is limited to the scenario of single text piece, while neglecting the studies at dialog level that should have more realistic values.
In this paper, we extend the ECPE task with a broader definition and scenario, presenting a new task, Emotion-Cause Quadruple Extraction in Dialogs (ECQED),
which requires detecting emotion-cause utterance pairs and emotion and cause types.
We present an ECQED model based on a structural and semantic heterogeneous graph as well as a parallel grid tagging scheme, which advances in effectively incorporating the dialog context structure, meanwhile solving the challenging overlapped quadruple issue.
Via experiments we show that introducing the fine-grained emotion and cause features evidently helps better dialog generation.
Also our proposed ECQED system shows exceptional superiority over baselines on both the emotion-cause quadruple or pair extraction tasks, meanwhile being highly efficient.
\end{abstract}


\section{Introduction}

\begin{figure}[!t]
\centering
\includegraphics[width=1\columnwidth]{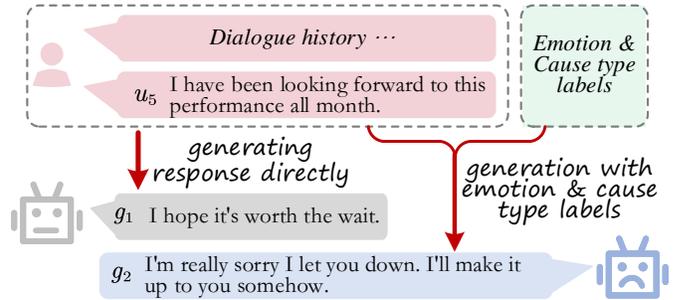}
\caption{
Illustration of the ECQED task (a). Refer to Table \ref{tab:statistics of types} and Section \ref{ssec:grid-tagging} for the details about the cause and emotion types. During dialog generation (b), with additional detailed emotion\&cause type label information, the response is much more empathic.
}
\label{fig:example1}
\end{figure}

\IEEEPARstart{I}t is a long-reached consensus in opinion mining area that emotion and cause are mutually indicative.
Emotion-Cause Pair Extraction task \cite{DBLP:conf/acl/XiaD19,DBLP:conf/acl/WeiZM20,DBLP:journals/taslp/ChengJYLG21,DBLP:journals/kbs/ChenSYH22} has accordingly been proposed to extract the emotion clauses and the corresponding cause clauses.
By revealing the underlying emotional causes, ECPE shows great potential on realist applications, such mental health analysis \cite{DBLP:conf/hci/TanLQM22,DBLP:conf/bibm/VajreNKS21} and public opinion monitoring \cite{DBLP:conf/emnlp/MiaoLL20,DBLP:journals/snam/KaramouzasMP22}.
Despite its promising success, the current ECPE research has two vital limitations, which unfortunately hinders its further prosperity and promotion.

From the task scenario perspective, existing studies only consider the single text piece (e.g., news articles or short posts), while another widely-occurred and sophisticated scenario, dialog, has not been studied in depth.
Although Emotion Recognition in Conversation (ERC) \cite{DBLP:conf/coling/LiJLZL20, DBLP:journals/taslp/Lian0T21, DBLP:conf/acl/ShenWYQ20} has been previously proposed to identify the speaker emotions in conversational utterance,
it suffers from the coarse-level analytics, i.e., without detecting the underlying causes of the emotions \cite{DBLP:conf/emnlp/GaoLDWCDX21, 9961847}.

From the task definition perspective, ECPE task only coarse-grainedly recognize which clauses (or utterances) are emotions or causes,
while ignoring the \emph{emotion and cause types}, the meta-information that reflects the finer grained opinion status.
Essentially, emotion type reveals what specifically the emotion is, and cause type depicts how the cause is aroused.
Intuitively, without leveraging such details, the utility of emotion-cause pairs (from ECPE) for the downstream applications (e.g., chatbots and empathic dialog generation) can be much limited \cite{DBLP:conf/naacl/SamadMFE22,DBLP:conf/emnlp/ShenZOZZ21}. 
As exemplified in Figure~\ref{fig:example1}(b), a dialog system manages to accurately generate a strong empathic response $g_2$ to the input utterance when the emotion and cause type label features are given.
Otherwise, an under-emotional response $g_1$ will be generated without understanding the emotion and its underlying cause due to the absence of type information.

Based on the above considerations, we investigate a new task, namely \textbf{E}motion-\textbf{C}ause \textbf{Q}uadruple \textbf{E}xtraction in \textbf{D}ialogs (\textbf{ECQED}).
Technically, given a dialog with utterances, ECQED sets the goal to extract each emotion-cause pair as well as their corresponding type labels, i.e., a quadruple (\emph{emotion utterance}, \emph{cause utterance}, \emph{emotion type}, \emph{cause type}), as seen in Figure~\ref{fig:example1}(a).
We build up the benchmark for ECQED task based on the existing RECCON dataset \cite{DBLP:journals/cogcom/PoriaMHGBJHGRCG21}, 
which includes 1,106 dialogs with human-annotated emotion-
cause quadruples.


Our ECQED task challenges in three aspects.
\textbf{First}, a dialog contains not only dialog text content but also other information such as the speaker identity, the order of utterance and the response relations between utterances.
\textbf{Second}, the overlapping phenomena of emotion or cause utterances are much more frequent than those in the news text.
As shown in Figure~\ref{fig:example1}(a), the cause utterance $u_1$ is paired with two emotion utterances $u_1$ and $u_2$.
\textbf{Third}, ECQED task is inherently a multi-stage task, having more steps to complete quadruple extraction than those in ECPE task.
Therefore, more errors will be propagated if simply building pipeline models.

To this end, we first propose an end-to-end quadruple extraction model equipped with a \textbf{Structural and Semantic Heterogeneous Graph} (SSHG),
which effectively captures the complex context information in the dialog.
For the overlapping issue, we design an efficient \textbf{parallel grid tagging module} that extracts emotion-cause pairs for each kind of emotion simultaneously. 
In addition, instead of decomposing the quadruple extraction problem into multiple stages, our model adopts the \textbf{joint extraction} method to avoid error propagation.

We evaluate our method on the RECCON dataset. 
Experimental results indicate that our model significantly outperforms those of 7 state-of-the-art baselines \cite{DBLP:conf/acl/DingXY20,DBLP:conf/acl/FanYDGYX20,DBLP:conf/coling/ChenHLWZ20,DBLP:conf/acl/WeiZM20,DBLP:conf/emnlp/DingXY20,DBLP:journals/taslp/ChengJYLG21,DBLP:journals/kbs/ChenSYH22}, 
which are adapted from the ECPE task or the RECCON paper.
Further ablation experiments demonstrate that each component of our framework is essential.
Specifically, the F1 score increases by 9.21\% after applying SSHG in our model.
The experiment for only overlapped quadruple extraction show that the F1 score of our model is 6.35\% higher than that of the best baseline. 
Moreover, emotion-cause quadruple information can facilitate the generation of empathic dialog, and the ROUGE scores are improved by average 5 points.
To facilitate related research and reproduce our results, all our codes and metadata will be publicly available after publication.

\begin{figure*}[!htbp]
	\centering
	\includegraphics[scale=0.39]{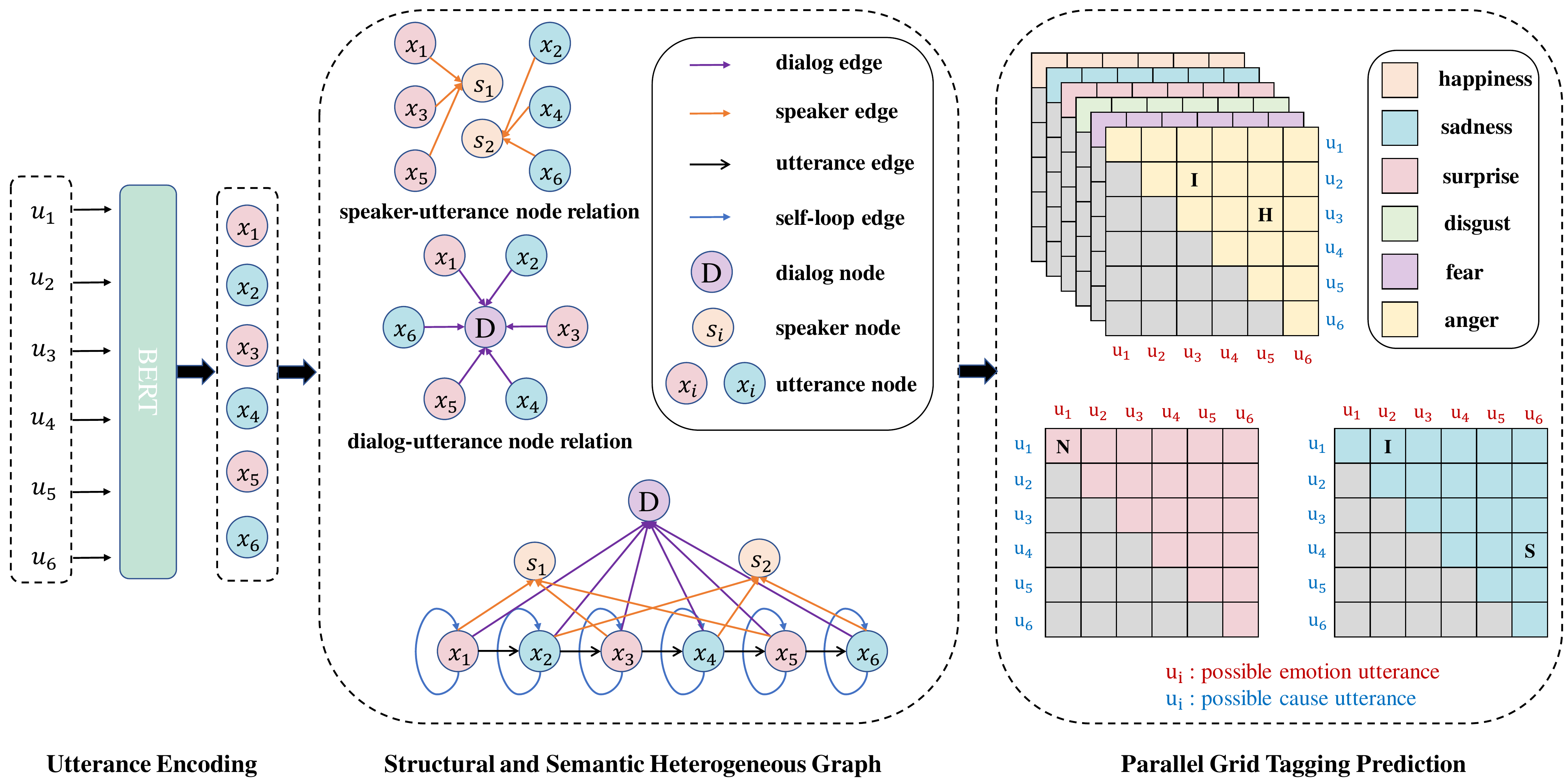}
	\vspace{-0.5\baselineskip}
	\caption{The overview of our model (best viewed in color). 
 In the graph, speaker nodes are initialized with pre-defined embeddings and dialog nodes are obtained by averaging the representations of all utterance embeddings. In the grid-tagging module, each emotion corresponds to a grid. And a quadruple can be identified by assigning a tag to the grid (cf. $\S$\ref{ssec:grid-tagging}).).
 }
	\label{fig:model}
\end{figure*}

\section{Related Work}
\subsection{Emotion Recognition in Conversation}
Emotion recognition in conversation (ERC) has attracted increasing attention in recent years \cite{DBLP:conf/acl/HuWH20,DBLP:journals/taslp/HsuSWC21,DBLP:conf/aaai/OngSCLNLSLW22}.
Different from traditional emotion recognition, emotion analysis in conversation is context-dependent and speaker-sensitive. 
It also needs to consider the speaker's state and dependency as well as a full understanding of the structure of the conversation.
Mainstream ERC methods are divided into two categories: sequence-based method \cite{DBLP:conf/coling/LiJLZL20,DBLP:conf/aaai/MajumderPHMGC19,DBLP:conf/emnlp/GhosalMGMP20} and graph-based method \cite{DBLP:conf/emnlp/GhosalMPCG19,DBLP:conf/emnlp/IshiwatariYMG20,DBLP:conf/acl/ShenWYQ20}.

The graph-based models better model the complicated conversational structures, comparing with the sequence-based methods. 
However, existing methods \cite{DBLP:conf/acl/ShenWYQ20,DBLP:conf/acl/HuLZJ20} focus on homogeneous graphs, while failing to utilize the vital interaction between speakers and utterances.
How to effectively leverage the heterogeneous graphs to model the dialog context and capture the interactions between speakers and utterances is still under-studied yet.

\subsection{Emotion-Cause Pair Extraction}
\label{ssec:ECPE}

Considering the fact that emotion and cause indicates each other,
Xia and Ding \cite{DBLP:conf/acl/XiaD19} proposed the ECPE task based on the Emotional Cause Extraction (ECE) task \cite{DBLP:conf/emnlp/GuiWXLZ16,DBLP:journals/access/YuRZOX19}.
The model is a two-step pipeline that first extracts emotion and cause clause, and then pairs them. 
Unfortunately, it leads to drawbacks of error propagation and high computational cost.
Later, end-to-end ECPE methods have been widely investigated \cite{DBLP:conf/acl/WeiZM20,DBLP:journals/corr/abs-2002-10710,DBLP:conf/acl/FanYDGYX20,DBLP:journals/taslp/ChengJYLG21,DBLP:journals/kbs/ChenSYH22}.
Wang et al. \cite{DBLP:journals/corr/abs-2110-08020} proposed multimodal emotion-cause pair extraction in conversations, and established a two-step pipeline model.
Because multimodal emotion analysis is not the research range in this study, we leave it in the future work.
However, all these existing ECPE works leave out the significance of emotion and cause types, and the neglect of them would result in quite limited performance as well as the downstream applications.

\subsection{The RECCON Dataset}
\label{ssec:RECCON}

To the best of our knowledge, the only work that is closely related to our ECQED task was performed by Poria et al. \cite{DBLP:journals/cogcom/PoriaMHGBJHGRCG21}. They built the RECCON dataset with human-annotated emotion-cause quadruples.
However, only two tasks, ``Causal Span Extraction'' and ``Causal Emotion Entailment'' were studies in their paper. The former is to identify the causal span given a target emotion utterance, while the latter is to identify the causal utterance that is similar with the ECPE task \cite{DBLP:conf/acl/XiaD19}.
By contrast, we in this paper extend their work based on their dataset by building up a strong model to jointly extract emotion-cause  quadruples (i.e., $<$emotion utterance, cause utterance, emotion type, cause type$>$) and carrying out extensive experiments to show the effectiveness of our model and the necessity of our ECQED task.

\section{Methodology}
\label{method}

In this paper, we propose an end-to-end quadruple extraction model to address this new ECQED task.
The architecture of our model is illustrated in Figure~\ref{fig:model}, which consists of three components.
First, we choose BERT \cite{DBLP:conf/naacl/DevlinCLT19}, the widely-used pre-trained language model, as encoder to yield contextualized utterance representations from input dialogs. 
Then, we design a SSHG to capture the structural and semantic information in dialogs and the interaction between utterances.
Finally, we adopt a grid tagging module to jointly reason the relationships between all pairs of utterances and extract the quadruples in parallel.

\subsection{Task Definition}
\label{ssec:uencode}

Let $U=\{u_1, u_2, ..., u_N\}$ be a dialog, where $N$ is the number of utterances.  
Each utterance $u_i=\{w_{i_1},…,w_{i_M}\}$ is a sequence of words of length $M$.
The goal of our task is to extract a set of quadruples in $U$:
\begin{equation} \label{eq:1}
P = \{...,(u_{e_j},u_{c_j},t_{e_j},t_{c_j}),...\}
\end{equation}
where $j$ means the j-th quadruple, $u_{e_j}$, $u_{c_j}$, $t_{e_j}$, $t_{c_j}$ denote the emotion utterance, cause utterance, emotion type and cause type, respectively.

\subsection{Utterance Encoding}
\label{ssec:uencode}

BERT \cite{DBLP:conf/naacl/DevlinCLT19} has achieved state-of-the-art performance across a variety of NLP tasks \cite{DBLP:conf/acl/WeiZM20,DBLP:conf/coling/ChenSLW0LJ22} due to its powerful text representation capacity.
Thus, we take BERT as the underlying encoder to yield contextualized utterance representations.
Concretely, for a given dialog $U=\{u_1, u_2, ..., u_N\}$, where the i-th utterance $u_i=\{w_{i_1}, w_{i_2}, …, w_{i_M}\}$, we insert a [CLS] token before each utterance and attach a [SEP] token to it, thus obtaining $u_i = ([CLS], w_{i_1} , w_{i_2} ,..., w_{i_M} , [SEP])$. 
Then we concatenate them together as the input of BERT to generate contextualized token representations, in which we take the representation of [CLS] token in each utterance $u_i$ as its utterance representation. 
After that, we obtain all the utterance representations $X = \{x_1, x_2,.., x_N\}$.

\subsection{Structural and Semantic Heterogeneous Graph}
\label{ssec:dgraph}

We design a structural and semantic heterogeneous graph to capture speaker, utterance and dialog features as well as the interaction between them. Formally,
we employ three distinct types of nodes in the graph:

\begin{itemize}
\item \textbf{Utterance nodes} summarize the information for each utterance and their context, which are obtained from the utterance encoder. 
\item \textbf{Dialog nodes} generalize the overall information of the dialog, which are initialized by averaging the representation of all utterance embeddings.
\item \textbf{Speaker nodes} represent the personality features of the speaker in the dialog. These nodes are obtained from pre-defined embeddings. 
\end{itemize}

The graph also has four types of edges:
\begin{itemize}
\item \textbf{Dialog-Utterance edges} connect the dialog nodes and utterance nodes, so that local and global context information can be exchanged.

\item \textbf{Speaker-Utterance edges} connect the spe-aker nodes and utterance nodes issued by the speaker, in order to model both the intra- and inter-speaker semantic interactions.

\item \textbf{Utterance-Utterance edges} connect the adjacent utterances in the chronological order.

\item \textbf{Self-loop edges} connect to the node itself, because emotion and cause utterances can be the same one in our task.

\end{itemize}

Motivated by R-GCN \cite{DBLP:conf/esws/SchlichtkrullKB18}, we apply a 2-layer graph convolution network to aggregate each node feature from its neighbors. 
The graph convolution operation is formulated as: 
\begin{equation} \label{eq:4}
h^{(l+1)}_n = ReLU(\sum_{k\in\kappa}\sum_{v\in N_k(n)}(W^{(l)}_k{h}^{(l)}_n + b^{(l)}_k))
\end{equation}
where $\kappa$ are different types of edges, $N_k(n)$ denotes the neighbors of the node $n$ connected with the $k^{th}$ type of edge, $W^{(l)}_k\in R^{d\times d}$ and $b^{(l)}_k\in R^d$ are the parameters of the $l^{th}$ layer.

\subsection{Parallel Grid Tagging Prediction}
\label{ssec:grid-tagging}

To extract quadruples, we first construct a grid for each emotion type and use four tags $\mathbb{T}$=\{H, I, N, S\} for each grid to represent the cause type for the emotion, which are explained as below:
\begin{itemize}
\item \textbf{H}ybrid: the cause lies in the joint influence of the speaker himself and other speakers. 
\item \textbf{I}nter-personal: the cause is in the utterance issued by another speaker.
\item \textbf{N}o-context: the cause exists in the utterance itself.
\item \textbf{S}elf-contagion: the cause is influenced by the speaker himself.
\end{itemize}

In order to further illustrate the process of grid tagging prediction, we present the grid-tagging module in Figure~\ref{fig:model} corresponding to the example in Figure~\ref{fig:example1}.
Due to the particularity of dialogs, the current utterance is only affected by the previous utterance or itself. 
Therefore, the cause utterance appear in front of the emotion utterance, which means that we can use the upper triangular grid. 
For example, as seen in Figure~\ref{fig:model}, in the grid corresponding to the emotion ``sadness'', S means that $u_4$ and $u_6$ are one emotion-cause pair and the emotion type is ``sadness'' and the cause type is ``self-contagion''.

Since the goal of our framework is to predict the relations (tags) between utterance pairs, it is crucial to generate a high-quality representation for the utterance pair grid.
We regard the representation $V_{ij}$  as a 3-dimensional matrix $V\in R^{N\times N\times d_h}$ of utterance pair ($u_i$,$u_j$).
Inspired by Lee et al. \cite{DBLP:conf/acl/LeeTXZ20}, we adopt the Conditional Layer Normalization (CLN)  to generate $V_{ij}$:
\begin{equation} \label{eq:4}
V_{ij}= CLN(h_i, h_j) =  \gamma_{ij} \odot N(h_j) + \lambda_{ij}
\end{equation}
where $N(h_j)=\frac{h_j - \mu}{\sigma}$, $\gamma_{ij} = W_\alpha h_i + b_\alpha$, bias $\lambda_{ij} = W_\beta h_i + b_\beta$, $\mu$ and $\sigma$ are the mean and standard deviation across the elements of $h_j$.

\noindent \textbf{MLP Predictor.} 
Based on the utterance pair grid representation $V$, we design a multi-layer perceptron (MLP) to calculate the relation (tag) scores for the utterance pair ($u_i$,$u_j$):\footnote{For simplicity, we only show the process of calculating tag scores for one emotion grid.}
\begin{equation} \label{eq:8}
y'_{ij} = MLP(V_{ij})
\end{equation}
where $y'_{ij} \in R^{|\mathbb{T}|}$ are the scores of the four pre-defined tags in $\mathbb{T}$. 
Prior work \cite{DBLP:conf/acl/LiXLFRJ21} has shown that MLP predictor can be enhanced by collaborating with a biaffine predictor. 
We thus follow such method and also add a biaffine predictor.

\noindent \textbf{Biaffine Predictor.} The inputs of the biaffine predictor are the outputs from the structural and semantic graph, i.e., $H = \{h_1, h_2, ..., h_N\}\in  R^{N\times d_h}$. 
Given a pair of utterance representations $h_i$ and $h_j$, we map them to an emotion-space representation $e_i$ and a cause-space representation $c_j$ with two MLPs, respectively.
Then, a biaffine classifier \cite{DBLP:conf/iclr/DozatM17} is used to calculate the relation scores for the utterance pair ($u_i$,$u_j$):
\begin{equation} \label{eq:5}
y''_{ij} = e_i^TUc_j + W[e_i;c_j] + b
\end{equation}
where $U$, $W$ and $b$ are trainable parameters.
The final relation probabilities $y_{ij}$ for the utterance pair ($u_i$,$u_j$) are calculated by combining the scores from the biaffine and MLP predictors:
\begin{equation} \label{eq:6}
y_{ij} = Softmax(y'_{ij} + y''_{ij})
\end{equation}

\subsection{Model Training}
\label{ssec:training}

The training loss is defined as the cross-entropy loss between the ground-truth tag distribution $\hat{y}_{ij}$ and the predicted tag distribution $y_{ij}$ for each utterance pair ($u_i$,$u_j$) in the emotion grid $e$:
\begin{equation} \label{eq:7}
\mathcal{L} = -\frac{1}{N^2} \sum_{e\in E}\sum_{i=1}^{n} \sum_{j=1}^{n}\sum_{k=1}^{|\mathbb{T}|}\hat{y}^{(e)}_{ij_k}log(y^{(e)}_{ij_k})
\end{equation}
where $E$ represents all the emotion grids.

\section{Experiments}
\label{experiment}

\subsection{Dataset, Implementation Details and Evaluation Metrics}
\label{ssec:data}

\begin{table}[t]
    \caption{Statistics of the numbers of conversations, utterances and quadruples in the RECCON dataset.}
  \centering
  \scalebox{1.2}{
    \begin{tabular}{c|ccc}
    \hline
     \textbf{RECCON} & \textbf{Train}&\textbf{Val} & \textbf{Test}   \\
    \hline
    Conversation & 834 & 47 & 225 \\
    Uterrances  & 8,206 & 493 & 2,405\\
    Quadruples & 7,029 & 328 & 1,767 \\
   \hline
    \end{tabular}%
    }

  \label{tab:dataset}%
\end{table}%

\begin{table}[t]
    \caption{Statistics of the numbers of emotion types and cause types in the RECCON dataset.}
  \centering
\resizebox{1.0\columnwidth}{!}{
    \begin{tabular}{lc}
    \hline
     \textbf{Emotion type} & \textbf{Number}   \\
    \hline
    happiness & 6,784  \\
    anger & 838 \\
    surprise & 644  \\
    sadness & 541  \\
    disgust & 220 \\
    fear & 97 \\
   \hline
    \end{tabular}%
    \hspace{2pt}
    \begin{tabular}{lc}
    \hline
     \textbf{Cause type} & \textbf{Number}   \\
    \hline
     no-context & 4,153 \\
     inter-personal & 2,548\\
     hybrid & 1,607 \\
    self-contagion & 816 \\
   \hline
    \end{tabular}%
    }

  \label{tab:statistics of types}%
\end{table}%

In this paper, we evaluate our model on the RECCON dataset \cite{DBLP:journals/cogcom/PoriaMHGBJHGRCG21}.
The statistics of the dataset are shown in Tables~\ref{tab:dataset} and \ref{tab:statistics of types}.
As seen, it consists of 1,106 dialogs and 11,104 utterances.
Each dialog contains multiple emotion-cause quadruples, and the average length of each dialog is 10 utterances, and the average number of emotion-cause quadruples in a dialog is 8. 
The dataset splitting follows the same setting in the experiments of the RECCON paper, with train:val:test as 7:1:2.

\begin{figure*}[!htbp]
	\centering
	\includegraphics[scale=0.475]{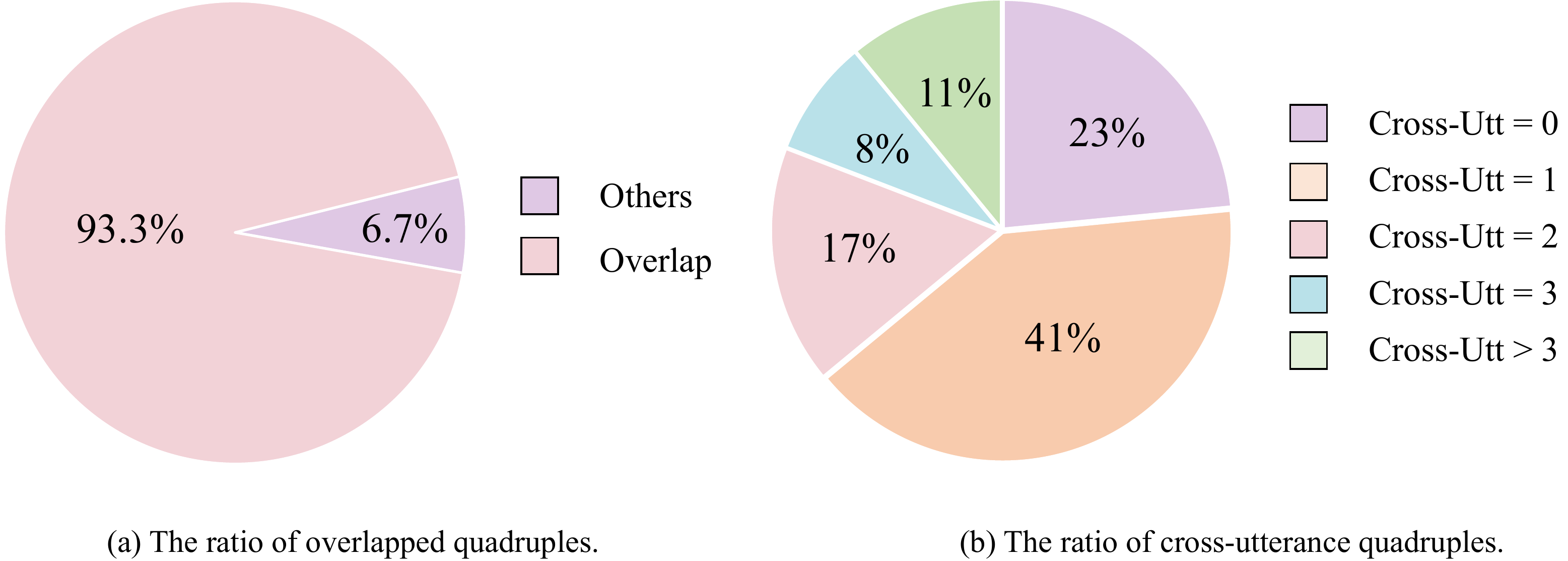}
	\vspace{-0.5\baselineskip}
	\caption{The ratio of overlapped quadruples (a) and cross-utterance quadruples (b). We define the distance between a pair of cause and emotion utterances in a quadruple as the number of utterances in-between them. ``0'' means that cause and emotion are in the same utterance.
 }
	\label{fig:cross}
\end{figure*}

In Figure~\ref{fig:cross}, we show the statistics of the ratios of overlapped quadruples and cross-utterance quadruples.
As shown in Figure~\ref{fig:cross}(a), we can observe that 93.3\% of the dialogs contain overlapped quadruples, which is much higher than the percentage in the ECPE dataset \cite{DBLP:conf/acl/XiaD19}.
As illustrated in Figure~\ref{fig:cross}(b),
a large proportion of quadruples in our dataset are not adjacent (36\%, where distance$>$1), and
the distance of 11\% of the quadruples is more than three utterance. 
Intuitively, the farther emotion utterance and cause utterance in a quadruple are, the more difficult they are to extract.
Therefore, considering the high ratios of overlapped and cross-utterance quadruples, our task is more challenging than other related tasks such as ECPE.

In Table~\ref{tab:hyperparameter}, we summarize the hyper-parameter setting in our model and experiments.
During training, we choose the Adam \cite{DBLP:journals/corr/KingmaB14} optimizer with the learning rate of 2e-5 for BERT \cite{DBLP:conf/naacl/DevlinCLT19} and 1e-5 for other modules.
We train our model by setting the epoch, dropout and batch size to 50, 0.2 and 2 respectively. 
Moreover, the hidden size $d_h$ is set as 768 to accommodate with BERT and the number of GCN layers is set to 2.
In terms of evaluation metrics, we follow previous ECPE works \cite{DBLP:conf/acl/XiaD19,DBLP:conf/acl/FanYDGYX20,DBLP:journals/taslp/ChengJYLG21,DBLP:journals/kbs/ChenSYH22} and use the Precision, Recall, and F1 score as the metrics for evaluation.
All our scores are the averaged number over five runs with different random seeds.

\begin{table}[t]
\caption{Details of the hyper-parameter settings.
}
  \centering
  \scalebox{1.25}{
    \begin{tabular}{c|c}
    \hline
     \textbf{Hyper-Parameters} & \textbf{Values}  \\
    \hline
    Leaning rate (BERT)  & 2e-5  \\
     Leaning rate (Other) & 1e-5  \\
     Batch size & 2 (dialogs) \\
     Epoch size & 50  \\
     Hidden size & 768 \\
     Layers of GCN & 2 \\
     Dropout & 0.2  \\
   \hline
    \end{tabular}%
    }
  \label{tab:hyperparameter}%
\end{table}%

\begin{table}[t]
    \caption{Results on quadruple extraction (\text{ECQED}). In the brackets are the improvements of our model over the best-performing baseline(s).
    }
  \centering
\fontsize{9}{11.5}\selectfont
\resizebox{0.76\columnwidth}{!}{
    \begin{tabular}{lccc}
    \hline
     & P     & R     & F1 \\
    \hline
    TransECPE & 53.46 & 58.12 & 55.69 \\
    ECPE-2D & 54.61 & 59.37 & 56.89 \\
    PairGCN & 54.92 & 59.71 & 57.21 \\
    RANKCP & 56.22 & 61.12 & 58.57 \\
    MLL & 54.01 & 65.25 & 59.11 \\
    UTOS & 52.63 & \textbf{68.20} & 57.27 \\
    RSN & 54.47 & 57.98 & 56.06 \\
 \textbf{Ours}  & \textbf{61.14} & 66.47 & \textbf{63.68} \\
 & \scriptsize{(+4.92\%})  & \scriptsize{(-1.73\%})  & \scriptsize{(+4.57\%}) \\
   \hline
    \end{tabular}%
    }

  \label{tab:quadruple extraction}%
\end{table}%

\begin{table*}[!t]
\caption{Experimental results on 1) emotion-cause pair extraction, 2) emotion utterance extraction and 3) cause utterance extraction.
    In the brackets are the improvements of our model over the best-performing baseline(s).
    }
  \centering
\fontsize{9}{11.5}\selectfont
\setlength{\tabcolsep}{2.4mm}
\resizebox{0.91\textwidth}{!}{
    \begin{tabular}{lccccccccc}
    \hline
   \multicolumn{1}{c}{\multirow{2}{*}{\textbf{Model}}}& \multicolumn{3}{c}{\textbf{Pair Extraction}} & \multicolumn{3}{c}{\textbf{Emotion Extraction}} & \multicolumn{3}{c}{\textbf{Cause Extraction}} \\
    
    \cline{2-4}\cline{5-7}\cline{8-10}
   
     & P     & R     & F1    & P     & R     & F1    & P    & R & F1\\
    \hline
    TransECPE & 59.32 & 64.49 & 61.79 & 80.39 & 87.41 & 83.75 & 76.33 & 82.99 & 79.52 \\
    ECPE-2D & 60.13 & 65.37 & 62.63 & 80.99 & 88.06 & 84.37 & 76.53 & 83.20 & 79.73\\
    PairGCN & 60.92 & 66.24 & 63.46 & 82.17 & 89.33 & 85.59 & 77.49 & 84.25 & 80.72\\
    RANKCP & 61.56 & 66.93 & 64.13 & 85.12 & 92.54 & 88.68 & 79.91 & 86.87 & 83.24\\
    MLL   & 61.97 & 67.38 & 64.56 & 78.93 & \textbf{95.34} & 86.36 & 77.97 & 84.77 & 81.23\\
    UTOS  & 61.83 & 67.22 & 64.41 & 78.63 & 94.98 & 86.03 & 73.99 & \textbf{89.38} & 80.96\\
    RSN   & 61.62 & 67.01 & 64.20 & 81.18 & 88.26 & 84.57 & 78.98 & 85.87 & 82.28\\
  \textbf{Ours}  & \textbf{66.59} & \textbf{72.39} & \textbf{69.37} & \textbf{87.66} & 95.31 & \textbf{91.32} & \textbf{81.91} & 89.05 & \textbf{85.48}\\
 & \scriptsize{(+4.62\%})  & \scriptsize{(+5.01\%})  & \scriptsize{(+4.81\%}) & \scriptsize{(+2.54\%})  & \scriptsize{(-0.03\%})  & \scriptsize{(+2.64\%}) & \scriptsize{(+2.00\%})  & \scriptsize{(-0.33\%})  & \scriptsize{(+2.24\%}) \\
    \hline
    \end{tabular}%
    }
  \label{tab: ECPE}%
\end{table*}%

\subsection{Baseline Systems}
\label{ssec:eval}

Since neither the RECCON paper nor prior work has implemented a model for our ECQED task, we adapted some state-of-the-art models for the ECPE task as our baselines because it shares some similarities with our task. 
Note that ECPE-2D, RANKCP and MLL were also used in the RECCON paper for their task.

\begin{itemize}

\item \textbf{ECPE-2D:} Ding et al. \cite{DBLP:conf/acl/DingXY20} proposed a two-dimensional expression scheme to recognize the emotion-cause pair.
\item \textbf{TransECPE:} Fan et al. \cite {DBLP:conf/acl/FanYDGYX20} proposed a transition-based model, which transforms the task into an analytic process of constructing a directed graph.
\item \textbf{PairGCN:} Chen et al. \cite {DBLP:conf/coling/ChenHLWZ20} designed a graph convolution network to model three kinds of dependencies between local neighborhood candidate pairs.
\item \textbf{RANKCP:} Wei et al. \cite{DBLP:conf/acl/WeiZM20} extracted emotion-cause pair from the perspective of ranking and through the graph attention network.
\item \textbf{MLL:} Ding et al. \cite{DBLP:conf/emnlp/DingXY20} transformed the ECPE task into the emotion-pivot cause extraction problem in the sliding window.
\item \textbf{UTOS:} Cheng et al. \cite{DBLP:journals/taslp/ChengJYLG21} reframed the emotion-cause pair extraction task as a unified sequence labeling problem.
\item \textbf{RSN:} Chen et al. \cite{DBLP:journals/kbs/ChenSYH22} jointly extracted emotion clauses, cause clauses and emotion-cause pairs through multi-task learning.

\end{itemize}

\subsection{Quadruple Extraction Results}
\label{ssec:eval}

The experimental results of emotion-cause quadruple extraction in terms of P, R and F1 measures are given in Table~\ref{tab:quadruple extraction}.
The results illustrate that our end-to-end method has obvious advantages over other strong baselines in the quadruple extraction task.
For example, our system surpasses the best baseline (MLL) with absolute 4.57\% F1 score.
Further analyzing, this performance gain mainly comes from the improvement on the precision score.
Compared with RANKCP, our precision is increased by 4.92\%, which proves that our model can effectively extract more correct quadruples.
Our remarkable performance reveals that the ECQED task is solvable, and meanwhile our approach is a feasible way to handle this task.

\subsection{Pair Extraction Results} 
\label{ssec:other task}

In Table~\ref{tab: ECPE} we compare our method with the existing methods of emotion-cause pair extraction to study the scalability of our model.
To ensure fairness, all models use BERT \cite{DBLP:conf/naacl/DevlinCLT19} as the encoder and are evaluated on the RECCON dataset.
The emotion-cause pair extraction results show that the F1 score of our model is 69.37\%, which is significantly better than those of other models.
Furthermore, in terms of emotion extraction and cause extraction, our model still shows competitive performance with the highest F1 scores 91.32\% and 85.48\%, respectively.
As a summary, the above experimental results confirm the well scalability of our model performing on other subtasks, though our model is initially proposed aiming at emotion-cause quadruple extraction.

\begin{table}[!t]
\caption{Ablation results on \text{ECQED} task. DU: Dialog-Utterance, SU: Speaker-Utterance.
    }
  \centering
\fontsize{9}{11.5}\selectfont
\resizebox{0.88\columnwidth}{!}{
    \begin{tabular}{lccc}
    \hline
    \textbf{} & P     & R     & F1 \\
    \hline
    \textbf{Ours}   & \textbf{61.14} & \textbf{66.47} & \textbf{63.68} \\    
    \quad -SSHG & 52.29 & 56.85 & 54.47 \\
    \quad -SU edge & 59.94 & 65.17 & 62.44 \\
    \quad -DU edge & 58.63 & 64.19 & 61.28 \\ 
    \quad -DU and SU edge & 58.75 & 60.25 & 59.49 \\
    \quad -MLP & 58.21 & 63.28 & 60.63 \\
    \quad -Biaffine & 59.83 & 65.05 & 62.33 \\ 
   \hline
    \end{tabular}%
    }
    
  \label{tab:different componets}%
\end{table}%

\subsection{Ablation Study}
\label{ssec:componets}

We conduct ablation experiments to further evaluate the contribution of each component.
As depicted in Table~\ref{tab:different componets}, we observe that no variants can compete with the complete model, suggesting that every component is essential for our task.
Specifically, the F1 score decreases most heavily without SSHG (-9.21\%), which indicates that it has a significant effect on modeling the complex structural information of dialog, such as the relationships between the context and speakers.
To verify the necessity and effectiveness of each node and edge of the heterogeneous graph, we remove the nodes and edges from the graph, respectively.
The sharp drops of results demonstrate that each node and edge plays an important role in capturing the speaker, utterance and dialog features as well as the interaction between them. 
Besides, removing MLP and biaffine predictors results in a distinct performance decline.
This suggests that the cooperation of two predictors is able to enhance our model performance.

\section{Analysis and Discussion}

\subsection{Do Fine-grained Emotion and Cause Features Really Help Empathic Dialog Generation?}

\begin{table}[t]
\caption{Experimental results of empathic dialog generation.`R' refers to the ROUGE metric.n. In the brackets are the improvements of the scheme 3) and 4) compared with the scheme 2), respectively.
    }
  \centering
\resizebox{1\columnwidth}{!}{
    \begin{tabular}{lcccc}
    \hline
     \textbf{Scheme} & R-1    & R-2     & R-L & Acc. \\
    \hline
    1) CU  & 0.4209 & 0.2450 & 0.4185 & 0.7615 \\
    2) CU+Pair & 0.4221 & 0.2461 & 0.4214 & 0.7889 \\
    \hline
    3) CU+ET  & 0.4369 & 0.2773 & 0.4517 & 0.7936 \\
    & \scriptsize{(+0.0148})  & \scriptsize{(+0.0312})  & \scriptsize{(+0.0303})  & \scriptsize{(+0.0047}) \\
    4) CU+ET+CT & \textbf{0.4673} & \textbf{0.3036} & \textbf{0.4650} & \textbf{0.8119} \\
     & \scriptsize{(+0.0452})  & \scriptsize{(+0.0575})  & \scriptsize{(+0.0436})  & \scriptsize{(+0.0230}) \\
 \hline
    \end{tabular}%
    }

  \label{tab:generation}%
\end{table}%

Holding the fact that fine-grained emotion and cause features can make generated utterances more empathetic, we set up the following experiments to comprehensively explore the effect of the ECQED task on the empathic dialog generation.
First of all, we extract each pair of adjacent utterances ($u_i$, $u_{i+1}$)  from the RECCON dataset where the former $u_i$ is the cause utterance and the latter $u_{i+1}$ is the emotion utterance.
We apply such data in the task of empathic dialog generation where cause utterance is used as input, and emotion utterance is used as the ground-truth of generated dialog.

We divide the experiments into 4 groups according to the features used for the dialog generation model.
1) \textbf{CU}: only using the \textbf{c}ause \textbf{u}tterance (the former one) as input into the dialog generation model.
2) \textbf{CU+Pair}: inputting the cause utterance and prompting that the next utterance is an emotion utterance but not giving the emotion type.
3) \textbf{CU+ET}: inputting the cause utterance and prompting the \textbf{e}motion \textbf{t}ype of the latter utterance.
4) \textbf{CU+ET+CT}: inputting the cause utterance and prompting the \textbf{c}ause \textbf{t}ype and emotion type.

We apply GPT-2 \cite{radford2019language} as our dialog generation model.
For measuring content quality, we use ROUGE metric \cite{lin-2004-rouge} to evaluate the fluency and relevance of generated utterances.
We fine-tune BERT \cite{DBLP:conf/naacl/DevlinCLT19} as the emotion classifier to predict the emotion label of the generated utterance, and use accuracy as an evaluation metric to verify the empathy performance of generative utterance.
As shown in Table~\ref{tab:generation}, the generation performance reaches the best when adding more fine-grained emotion and cause features, such as cause utterances, cause types and emotion types. 
For example, by incorporating both the cause types and emotion types, the ROUGE metric is increased by average 5 points, i.e., 4) vs. 2). 
In addition, the dialog generation model enhanced with CU+ET+CT features achieves the best accuracy, suggesting that it is able to generate more empathic utterances with correct emotion types.

\subsection{Special Comparison on Overlapped Quadruple Extraction}
\label{ssec:overlap}

\pgfplotsset{compat=newest}

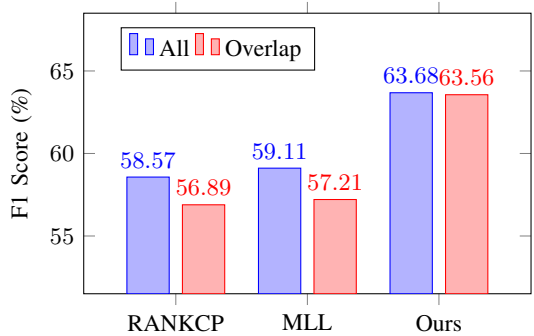
\begin{figure}[!t]
\centering
\begin{tikzpicture}
[font=\small]
\begin{axis}[
    ybar,
    height=0.6\columnwidth,
    width=0.85\columnwidth,
    enlargelimits=0.35,
    legend style={at={(0.3,0.95)},
      anchor=north,legend columns=-1},
    ylabel={F1 Score (\%)},
    symbolic x coords={RANKCP,MLL,Ours},
    xtick=data,
    ymin= 55, ymax= 65,
    ybar=5pt,
    bar width=16pt,
    nodes near coords,
    nodes near coords align={vertical},
    ]

\addplot coordinates {(RANKCP,58.57) (MLL,59.11) (Ours,63.68)};
\addplot coordinates {(RANKCP,56.89) (MLL,57.21) (Ours,63.56)};
\legend{All,Overlap}
\end{axis}
\end{tikzpicture}
\caption{Comparison with RANKCP and MLL on overlapped quadruple extraction. ``All'' and ``Overlap'' mean that the results are calculated based on all quadruples and overlapped quadruples solely.
}
\label{fig:overlap}
\end{figure}

We are curious about the impact of overlapped quadruples on model performance.
Therefore, we make up a subdataset (i.e., Overlap) that only containing overlapped quadruples in the test set.
In Figure~\ref{fig:overlap}, we compare our model with two competitive baselines on the Overlap and All datasets, i.e., RANKCP and MLL.
It can be seen that our model always performs better than MLL and RANKCP on extracting quadruples, no matter whether they are overlapped or not.
Notably, our model wins MLL over 6.35\%(63.56-57.21) F1 score.
What's more, the performance gaps are further enlarged when considering overlapped quadruple extraction.
The overall results reveal that our method can well solve the problem of overlapped quadruples.

\subsection{Advantage of Joint Emotion-Cause Quadruple Extraction}
\label{ssec:pip}

\begin{figure*}[!ht]
	\centering
	\includegraphics[scale=0.485]{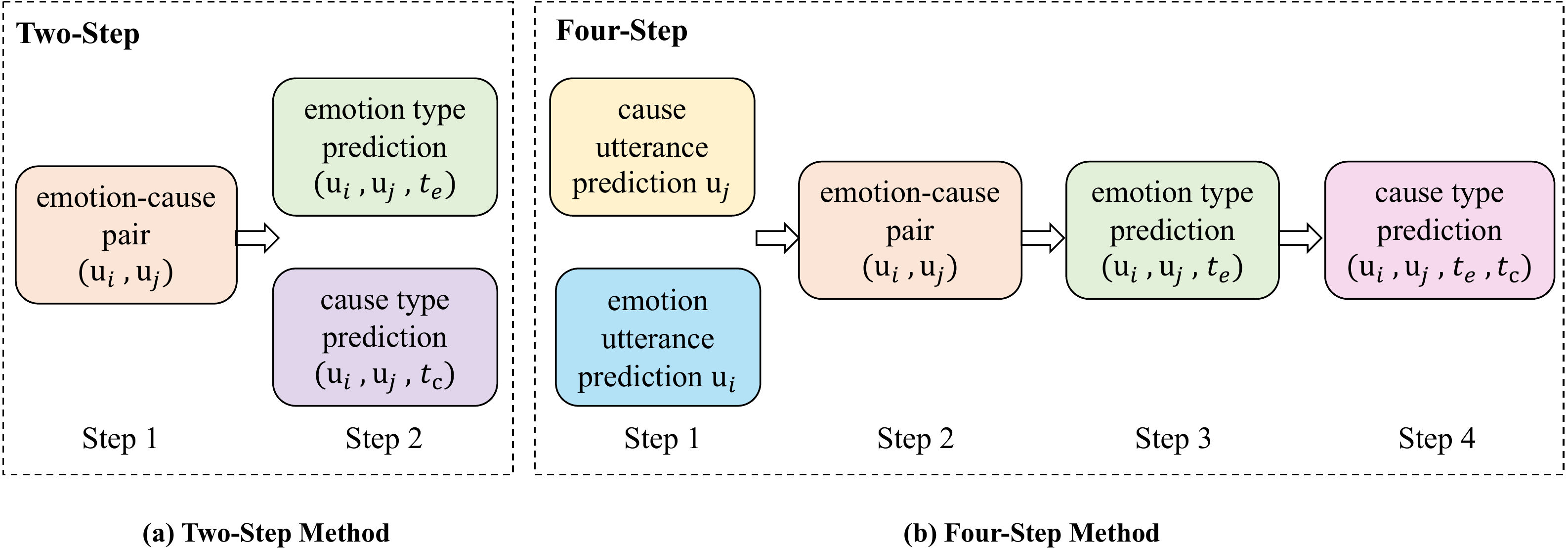}
	\vspace{-0.5\baselineskip}
	\caption{Flow chart of two pipeline methods for quadruple extraction.}
	\label{fig:pipeline}
\end{figure*}

In this part, we verify the effectiveness of our end-to-end method by designing two pipeline methods to compare with it.
The first method is ``Four-Step'' scheme for the quadruple extraction, whose framework is shown in Figure \ref{fig:pipeline}(a): 
1) extract emotion utterances and cause utterances;
2) perform emotion-cause pairing and filtering;
3) classify the emotion types of emotion-cause pairs;
4) classify the cause types of emotion-cause pairs.
In contrast, the second method consists of two steps, as illustrated in Figure \ref{fig:pipeline}(b): 
1) extract the emotion-cause pairs;
2) classify the emotion types and cause types of the extracted emotion-cause pairs.
As seen in Table \ref{tab:pipeline}, our end-to-end method achieves a 16.27\% 
(63.68-47.41) higher F1 score than the four-step method, and a 8.29\% (63.68-55.39) higher F1 score than the two-step method. 
This implies that our end-to-end method avoids error propagation.

\begin{table}[t]
\caption{Comparison with pipeline methods on \text{ECQED} task.}
\centering
\fontsize{10}{11.5}\selectfont
\scalebox{1.0}{
\begin{tabular}{lccc}
\hline
\textbf{Scheme} & P     & R     & F1 \\
\hline
Four-Step  & 45.49 & 49.47 & 47.41 \\
Two-Step & 53.18 & 57.81 & 55.39 \\
\bf Ours  & \textbf{61.14} & \textbf{66.47} & \textbf{63.68} \\
\hline
\end{tabular}%
}

\label{tab:pipeline}%
\end{table}%

\begin{table}[t]
\caption{Comparison of grid tagging using one grid and multiple grids. 
``All'' and ``Overlap'' represent the same meanings as mentioned in Figure~\ref{fig:overlap}.
}
\centering
\scalebox{1.2}{
\begin{tabular}{cccc}
\hline
\textbf{Scheme}&  F1 (All)    & F1 (Overlap)   & Speed (sent./s) \\
\hline
One-Grid  & 62.87 & 60.89 &  77.58\\
Multi-Grid & \textbf{63.68} & \textbf{63.56} &  \textbf{104.56}\\

\hline
\end{tabular}%
}
\label{tab:grid tag}%
\end{table}%

\subsection{Efficiency Study}

In order to further demonstrate the advantage of our parallel grid tagging module, we design a model using one grid to decode all kinds of quadruples to compare with our multi-grid model.
Most parts of one-grid model are the same
as the ones of multi-grid model, except the grid tagging layer.
As there is only one grid, the tags are designed as the combinations of emotion and cause types.
For example, if the emotion type is ``surprise'' and cause type is ``hybrid'', the tag should be ``SU-H''.
Besides performances, we also compare the efficiencies of two models by running these methods in the same hardware environment (i.e., NVIDIA RTX 3090 GPU).

As shown in Table \ref{tab:grid tag}, the F1 score for extracting all quadruples using the multi-grid method is 0.81\% higher than that of the one-grid method.
Particularly, the advantage of the multi-grid method for extracting overlapped quadruples is more obvious, and the F1 score is 2.67\% higher than the one-grid method. 
Apart from the model performance, the inference speed of the multi-grid method is about $\times$ 1.35 times faster than the one-grid method, which proves the efficiency of our parallel grid tagging module.


\subsection{Case Analysis}

\begin{table*}[!t]
\caption{Case Study to compare with the methods without ``Structural and Semantic Heterogeneous Graph'' and ``Parallel Grid Tagging Prediction''. The \textcolor{red}{red} elements mean wrong predictions and the \st{strikethrough} ones represent missing predictions. The expression of an emotion type uses its first two letters, e.g., surprise-SU.
}
\centering
\resizebox{1.0\textwidth}{!}{
\begin{tabular}{lccc}
\hline
\multicolumn{1}{c}{\textbf{Dialog Utterances}} & \textbf{-SSHG} & \textbf{-Parallel Prediction}  & \textbf{Ground-Truth}\\
\hline
\makecell[l]{[$\blacktriangleright u_1$] What? How could you forget to reserve the concert tickets? \\ $\lbrack\blacktriangleright u_2\rbrack$ I'm sorry. I forget all about it.
$\lbrack\blacktriangleright u_3\rbrack$ How could you? \\ I reminded you just this morning.
$\lbrack\blacktriangleright u_4\rbrack$ It's leap my mind. \\ I really feel terrible about it.
$\lbrack\blacktriangleright u_5\rbrack$ I have been looking forward \\ to this performance all month.
$\lbrack\blacktriangleright u_6\rbrack$ I'm really sorry I let you \\ down. I'll make it up to you somehow.}
  & \makecell[c]{\textcolor{red}{($u_1$,$u_1$, AG, N)}\\($u_2$, $u_1$, SA, I)\\($u_3$, $u_2$, AG, I)\\\textcolor{red}{($u_5$,$u_3$, AG, S)}\\($u_6$,$u_4$, SA,S)}   & 
  
  \makecell[c]{\sout{($u_1$,$u_1$, SU, N)}\\ ($u_2$,$u_1$, SA, I)\\ ($u_3$,$u_2$,AG, I)\\ \textcolor{red}{($u_5$,$u_4$, SA,I)} }   & 
  
  \makecell[c]{($u_1$,$u_1$, SU, N)\\($u_2$, $u_1$, SA, I)\\($u_3$, $u_2$, AG, I)\\($u_5$,$u_3$, AG, H)\\($u_6$,$u_4$, SA,S) }\\ 
\hline
\makecell[l]{$\lbrack\blacktriangleright u_1\rbrack$ This is the end, Jane. I don't want a girlfriend who \\ goes out with other guys all the time. $\lbrack\blacktriangleright u_2\rbrack$ I won't do it \\ again. Please forgive me. $\lbrack\blacktriangleright u_3\rbrack$ No way. I've given you too \\ many chances already.} & \makecell[c]{\textcolor{red}{($u_1$,$u_1$, SA, N) }\\ ($u_2$,$u_1$, SA, I) \\($u_3$,$u_1$, AG, S)} & \makecell[c]{\sout{($u_1$,$u_1$, SA, N)}\\($u_2$,$u_1$, SA, I)\\
\textcolor{red}{($u_3$, $u_2$, AG, I)} \\
}  & \makecell[c]{($u_1$,$u_1$, AG, N) \\ ($u_2$,$u_1$, SA, I) \\($u_3$,$u_1$, AG, S)}\\
\hline

\end{tabular}
}

\label{tab:case study} 
\end{table*}

We empirically perform case study on the ECQED task to better understand the capacity of our proposed model. 
Specifically, in Table~\ref{tab:case study}, we show some predictions for two instances on the test set.
Our complete model can correctly extract all quadruples in these instances. 
By contrast, the model incorrectly predicts the types of emotion or cause without SSHG. 
Through analysis, one possible reason is that the lack of reasoning ability between the context and speaker's state.
Meanwhile, when parallel prediction is not employed, for the case of overlapped quadruples, such as ($u_1$,$u_1$, SU, N) and ($u_2$, $u_1$, SA, I) in the first example, only the latter one is extracted. 
There is also a pairing error in each example, i.e., ($u_5$, $u_4$, SA, I) and ($u_3$, $u_2$, AG, I).
In summary, the above analyses indicate that SSHG plays a crucial role in modeling sophisticated dialog structures and semantic understanding. 
Moreover, the parallel prediction method is indispensable in solving the problem of overlapped quadruple extraction.

\section{Conclusion}

In this paper, we propose a novel task, Emotion-Cause Quadruple Extraction in Dialogs (ECQED), detecting the fine-grained quadruple: \emph{emotion utterance}, \emph{cause utterance}, \emph{emotion type}, \emph{cause type}.
We then present a nichetargeting model for ECQED, which leverages rich dialog context information, and meanwhile effectively addresses the severe quadruple overlapping problem and error-propagation problem in this task.
The experimental results show that our end-to-end model achieves new SOTA performance on the task,
and further ablation study evaluates the efficacy of our model design.
Moreover, we show that emotion-cause quadruples can facilitate empathic dialog generation, demonstrating the necessity of this task.
Our work pushes forward the research of emotion-cause analysis from both the scenario (i.e., dialog) and granularity (i.e., quadruple).

In future work, we suggest the following directions to extend our study:

\noindent$\blacktriangleright$ \textbf{Enhancing the generation of controllable empathic dialog}

In this paper, we have utilized emotion-cause type information as prompts to promote the generation of controllable empathic dialog, and obtained decent result improvement. 
In future research, one can further consider how to employ type information in a more complex way to further improve empathic dialog generation.

\noindent$\blacktriangleright$ \textbf{Extracting implicit emotion-cause quadruples}

Implicit emotion-cause extraction aims to detect the emotion or cause expressions given in a latent manner (e.g., $u_5$ in Figure \ref{fig:example1}), which is a much more challenging task.
Although some prior work has investigated implicit sentiment analysis \cite{li-etal-2021-learning-implicit}, the method using sentiment dictionary to distinguish explicit and implicit sentiment expressions may be not suitable for emotions, because emotions are more fine-grained than sentiments.
A possible solution is to build an emotion dictionary to distinguish explicit and implicit emotions first and then carry out nichetargeting research on implicit emotion-cause extraction.

\section*{Acknowledgments}

This work is supported by the National Natural Science Foundation of China (No. 62176187), the National Key Research and Development Program of China (No. 2022YFB3103602).

\bibliography{ref}
\bibliographystyle{IEEEtran}


\end{CJK}
\end{document}